\documentclass[iicol, sn-basic]{sn-jnl}

\usepackage{graphicx}%
\usepackage{multirow}%
\usepackage{amsmath,amsfonts,amssymb}
\usepackage{amsthm}%
\usepackage{mathrsfs}%
\usepackage[title]{appendix}%
\usepackage{xcolor}%
\usepackage{textcomp}%
\usepackage{manyfoot}%
\usepackage{booktabs}%
\usepackage{algorithm}%
\usepackage{algorithmicx}%
\usepackage{algpseudocode}%
\usepackage{listings}%

\usepackage{booktabs}
\usepackage{makecell}
\usepackage{amsmath,amssymb} 
\usepackage{pifont}
\newcommand{\cmark}{\ding{51}}%
\usepackage{multirow}

\newcolumntype{x}[1]{>{\centering\arraybackslash}p{#1pt}}
\newcommand{\tablestyle}{\setlength{\tabcolsep}{3pt}\renewcommand{\arraystretch}{1.0}\centering\footnotesize}
\newcommand{\scale}{\setlength{\tabcolsep}{10pt}\renewcommand{\arraystretch}{1.0}\centering\normalsize}

\usepackage{colortbl}
\newcommand{\gray}[0]{\cellcolor[gray]{0.9}}

\usepackage[capitalize]{cleveref}
\crefname{section}{Sec.}{Secs.}
\Crefname{section}{Section}{Sections}
\Crefname{table}{Table}{Tables}
\crefname{table}{Tab.}{Tabs.}
\usepackage{subfig}

\begin{document}

\title[Article Title]{Geometric-aware Pretraining 
for Vision-centric 3D Object Detection}


\author[]{\fnm{Linyan} \sur{Huang}$^{1,2}$}

\author[]{\fnm{Huijie} \sur{Wang}$^{2}$}

\author[]{\fnm{Jia} \sur{Zeng}$^{2,3}$}

\author*[]{\fnm{Shengchuan} \sur{Zhang$^{1*}$}}\email{zsc\_2016@xmu.edu.cn}

\author[]{\fnm{Liujuan} \sur{Cao}$^{1}$}

\author[]{\fnm{Junchi} \sur{Yan}$^{3}$}

\author[]{\fnm{Hongyang} \sur{Li}$^{2,3}$}

\affil[1]{
\orgname{Department of Artificial Intelligence, School of Informatics, Xiamen University}
}
\affil[2]{
\orgdiv{OpenDriveLab, Shanghai AI Lab .\quad $^{3}$ Shanghai Jiao Tong University}
}


\abstract{
Multi-camera 3D object detection for autonomous driving is a challenging problem that has garnered notable attention from both academia and industry. An obstacle encountered in vision-based techniques involves the precise extraction of geometry-conscious features from RGB images. Recent approaches have utilized geometric-aware image backbones pretrained on depth-relevant tasks to acquire spatial information. However, these approaches overlook the critical aspect of view transformation, resulting in inadequate performance due to the misalignment of spatial knowledge between the image backbone and view transformation. To address this issue, we propose a novel geometric-aware pretraining framework called \textbf{GAPretrain}. Our approach incorporates spatial and structural cues to camera networks by employing the geometric-rich modality as guidance during the pretraining phase. The transference of modal-specific attributes across different modalities is non-trivial, but we bridge this gap by using a unified bird's-eye-view (BEV) representation and structural hints derived from LiDAR point clouds to facilitate the pretraining process. GAPretrain serves as a plug-and-play solution that can be flexibly applied to multiple state-of-the-art detectors. Our experiments demonstrate the effectiveness and generalization ability of the proposed method. We achieve 46.2 mAP and 55.5 NDS on the nuScenes \textit{val} set using the BEVFormer method, with a gain of 2.7 and 2.1 points, respectively. We also conduct experiments on various image backbones and view transformations to validate the efficacy of our approach. Code will be released at \url{https://github.com/OpenDriveLab/BEVPerception-Survey-Recipe}.
}




\maketitle

\section{Introduction}\label{sec1}

\begin{figure}[t!]
    \centering
    \includegraphics[width=\linewidth]{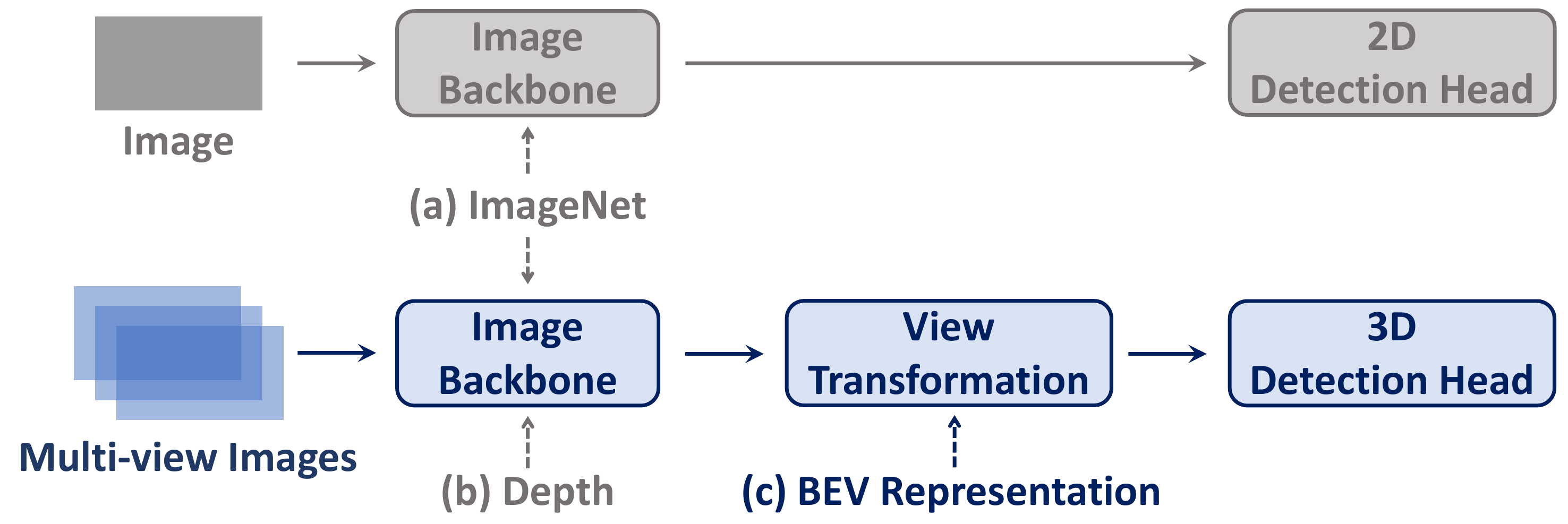}
    \caption{
    \textbf{Comparison on 
    different pretraining approaches.}
    (a) For the task of 2D detection, image backbones are pretrained on ImageNet for initialization.
    (b) In 3D detection, depth information plays a critical role. 
    The performance of pretrained backbones on depth-relevant tasks is superior to that on ImageNet. 
    (c) In view transformation, the module is initialized randomly, which means the spatial information encoded by the image backbone might not be fully utilized. 
    Our proposed pretraining approach offers initialized weights for both the image backbone and view transformation.
    }
    \vspace{-10pt}
    \label{fig:compare}
\end{figure}

Camera-based 3D object detection is a crucial vision task for perceiving the surrounding environment, and it has been extensively researched in the field of autonomous driving~\citep{tellex2011understanding, li2022Recipe, chen2022persformer, hu2022goal, hu2022st, wutrajectory, wupolicy}. And most of the recent state-of-the-art approaches employ bird's-eye-view (BEV) representation~\cite{li2022bevformer, li2022bevdepth}. In 2D detection tasks, the integration of conventional image backbones~\citep{he2016resnet, lee2019vovnet} is commonly employed to achieve promising performance. However, when it comes to BEV-based detection, the rich semantic information provided by image backbone alone is insufficient. This is due to the fact that accurate BEV feature representation heavily relies on the geometric cues, while mining 3D geometric cues from 2D images is a challenging and ill-posed problem. To address this issue, we propose a geometric-aware pretraining distillation method for BEV-based 3D object detectors that allow us to tap into the potential of image backbones.

As shown in Fig. \ref{fig:compare}, the BEV-based detection network consists of three components: an image backbone for feature extraction, a view transformation module for converting perspective view features into BEV, and a detection head for predicting 3D bounding boxes. Learning geometry-relevant representations from RGB images is the core and most challenging aspect of vision-based 3D detectors, in contrast to the 2D detection task. Such challenge poses high demands on the view transformation module.

Although image backbones pretrained on ImageNet provide semantically rich features, they fail to capture critical structural information. 
The existing pre-training method~\citep{park2021dd3d} introduces depth-relevant tasks, but such method excludes the view transformation module and benefits the backbone only. View transformation module is of great significance, since it constructs 3D information and encodes 3D prior assumptions. As the view transformation module is optimized only by the detection loss, the spatial information from the depth-pretrained backbone is not fully utilized. Ideally, the image backbone and the view transformation should incorporate aligned geometry knowledge to enhance their performance.

LiDAR point clouds, which contain rich spatial information, offer superior performance on 3D detection benchmarks compared to camera-based methods~\citep{sun2020scalability, caesar2020nuscenes}. Therefore, integrating LiDAR information into camera-based models is a natural step. Recent works~\citep{guo2021liga,peng2022lidar} have explored LiDAR information in two paradigms. The first paradigm~\citep{li2022bevdepth} projects LiDAR points onto the perspective view to provide additional depth supervision. However, the task of depth estimation in 2D space is inherently ill-posed and challenging. The second paradigm~\citep{chong2022monodistill,li2022uvtr,chen2022bevdistill} leverages a teacher-student knowledge distillation fashion. Nonetheless, it is suboptimal to transfer knowledge between these two modalities in an end-to-end manner, as the feature representations of these modalities hold distinctions. State-of-the-art distillation methods~\citep{chen2022bevdistill} are primarily focused on foreground-guided distillation and instance-wise distillation. Although it makes an attempt to address cross-modal domain adaptation issues, its effectiveness is limited. Hence, we propose the pretraining distillation paradigm to overcome the domain gap.

We propose a geometric-aware pretraining method called \textbf{GAPretrain}, which integrates geometric-rich features into camera-based 3D object detectors. Our contribution can be summarized as follows:
\textbf{(a)} We introduce a new way to leverage geometry-aware knowledge by utilizing the unified BEV representation to inject both geometric and semantic information into camera-based models, rather than constructing depth-relevant supervision for the image backbone.
\textbf{(b)} By jointly pretraining the image backbone and the view transformation, vision-centric object detectors can learn more accurate spatial information, resulting in enhanced model performance.
\textbf{(c)} To alleviate the misalignment between different modalities, we propose two new modules: the LiDAR-guided Mask Generation and the Target-aware Geometry Correlation. The LiDAR-guided Mask Generation module helps camera-based detectors concentrate on valuable representative regions and overlook noisy background areas. The Target-aware Geometry Correlation module focuses on object-level geometry information, such as orientation and scale, instead of pixel-wise knowledge transfer.
\textbf{(d)} We show that the proposed method serves as a plug-and-play recipe extensible to various state-of-the-art camera-based methods. Experimentally, we demonstrate that our approach universally improves performance, regardless of the network design. When built upon BEVFormer, our method yields a significant improvement in mAP and NDS of 2.7 and 2.1 points on the nuScenes \textit{val} set, respectively.

\section{Related Work}

\subsection{Visual Pretraining}

Pretraining has become a necessary requirement for achieving good performance in many 2D computer vision tasks. However, for the task of 3D perception, state-of-the-art models are typically trained from scratch. While some methods~\citep{zhang2021self,xie2020pointcontrast} have employed unsupervised or self-supervised paradigms to promote representation learning for networks consuming point cloud data, the pretraining method for models taking 2D images as input has not been thoroughly explored.
For camera-based methods~\citep{park2022time, zhang2022beverse, liu2022petr, huang2021bevdet, li2022bevformer, wang2022detr3d}, image backbones are typically pretrained on ImageNet~\citep{deng2009imagenet} or DDAD15M~\citep{park2021dd3d} to enhance the semantic or the depth-relevant representation. However, these backbones offer suboptimal solutions for perceiving spatial information. In this work, we shed light on pretraining methodology for camera models by proposing to incorporate supervision signals from another geometric-rich modality.

In recent years, researchers have explored various approaches for leveraging information from one modality to guide the pretraining process of another. One early proposal by ~\cite{gupta2016cross} first proposed to train a network on an unlabeled modality with the supervision of a learned model on another labeled modality. ~\cite{liu2021learning} introduced a learned 2D model to pretrain a 3D counterpart by contrastive learning between two modalities. This approach achieved promising results on an indoor point cloud dataset for detection and segmentation tasks. Another recent approach, SLidR~\citep{sautier2022image2lidar}, used a camera model to facilitate the feature learning of a LiDAR network through contrastive self-supervised learning. Meanwhile, Pri3D~\citep{hou2021pri3d} pretrained 2D segmentation networks with contrastive learning under geometric constraints from point cloud data. Although previous methods have focused on transferring knowledge from 2D models or using cross-modal contrastive learning, our work proposes to transfer knowledge from a geometric-aware representation to guide the learning of 2D networks, in a novel approach that we believe can help advance the field further.

\subsection{Knowledge Distillation}

Knowledge distillation~\citep{hinton2015distilling} is initially proposed to transfer the learned knowledge from a large network to a small one for model compression.
This strategy has been proven effective in various computer vision tasks, such as 2D object detection~\citep{chen2017learning, wang2019distilling, dai2021general, yang2022focal} as well as 3D domains~\citep{cho2022itkd, yang2022dis, zhang2022pointdistiller}.

Our work relates to transferring knowledge between two modalities via distillation.
MonoDistill~\citep{chong2022monodistill} considered from both feature and result space to guide the camera model.
LIGA-Stereo~\citep{guo2021liga} learned a stereo-based model to imitate the high-level geometry-aware representation from a LiDAR-based model in 3D and BEV space.
UVTR~\citep{li2022uvtr} transferred knowledge from geometry-rich teacher to geometry-inferior student by minimizing distance without excluding background features.
BEV-LGKD~\citep{li2022bev} used the LiDAR points as guidance, which ignored the misalignment between LiDAR and image data.
BEVDistill~\citep{chen2022bevdistill} utilized ground truth as the heatmap for pixel-wise distillation. It overlooked the global interaction, which is important for different modalities.
These works train the knowledge distillation framework in an end-to-end manner alongside the target task in the model, which may introduce difficulty in optimization due to the vast modality gap.
Our framework mitigate this problem by formulating a pretraining procedure to introduce spatial knowledge.
Besides, our method can fully utilize unlabeled data to further boost model performance.

\subsection{3D Object Detection}

\textbf{Camera-based 3D Object Detection.} 
Recent years have witnessed an extraordinary blossom of vision-centric approaches to achieve impressive results on many benchmarks. 
Due to several benefits of performing perception tasks in BEV space~\citep{li2022bevformer},
recent methods tend to conduct view transformation first to obtain BEV features, and then predict detection results based on the BEV features.

Lift-Splat-Shoot (LSS)~\citep{philion2020lss} proposed leveraging depth distribution to model uncertainty in depth estimation, and plenty of works follow this paradigm for perspective transformation~\citep{huang2021bevdet, li2022bevdepth}.
To further solve the issue of inaccurate depth estimation, BEVDepth~\citep{li2022bevdepth} utilized explicit depth supervision. 
BEVFormer~\citep{li2022bevformer} performed view transformation by exploiting deformable attention and devised grid-shaped queries in the BEV space.
In this work, we also aim for the camera-based detection task, and propose a novel pretraining scheme 
to improve the performance of existing state-of-the-art detectors.

\textbf{LiDAR- and Fusion-based 3D Object Detection.}
LiDAR-based approaches~\citep{yan2018second, yin2021centerpoint, zhou2018voxelnet, lang2019pointpillar} utilize accurate spatial information from point cloud data and thus yield superior performance compared to camera-based solutions. 
To further boost performance, fusion-based methods take LiDAR and camera data as input, leveraging both the geometry and semantic advantages in two modalities.
TransFusion~\citep{bai2022transfusion} adopted the bounding box prediction as a proposal to query the image feature, then fused the visual information to LiDAR features.
BEVFusion~\citep{liang2022bevfusion} proposed a simple yet effective pipeline to combine BEV representations from different sensors and achieved impressive results.
However, with regards to the deployment and computational cost of LiDAR sensors, many real-world autonomous vehicles resort to a camera-only configuration.
We leverage LiDAR signals only in the training phase, and add no extra expense for the inference of camera models.


\section{Method}

\begin{figure*}[t!]
    \centering
    \includegraphics[width=.95\linewidth]{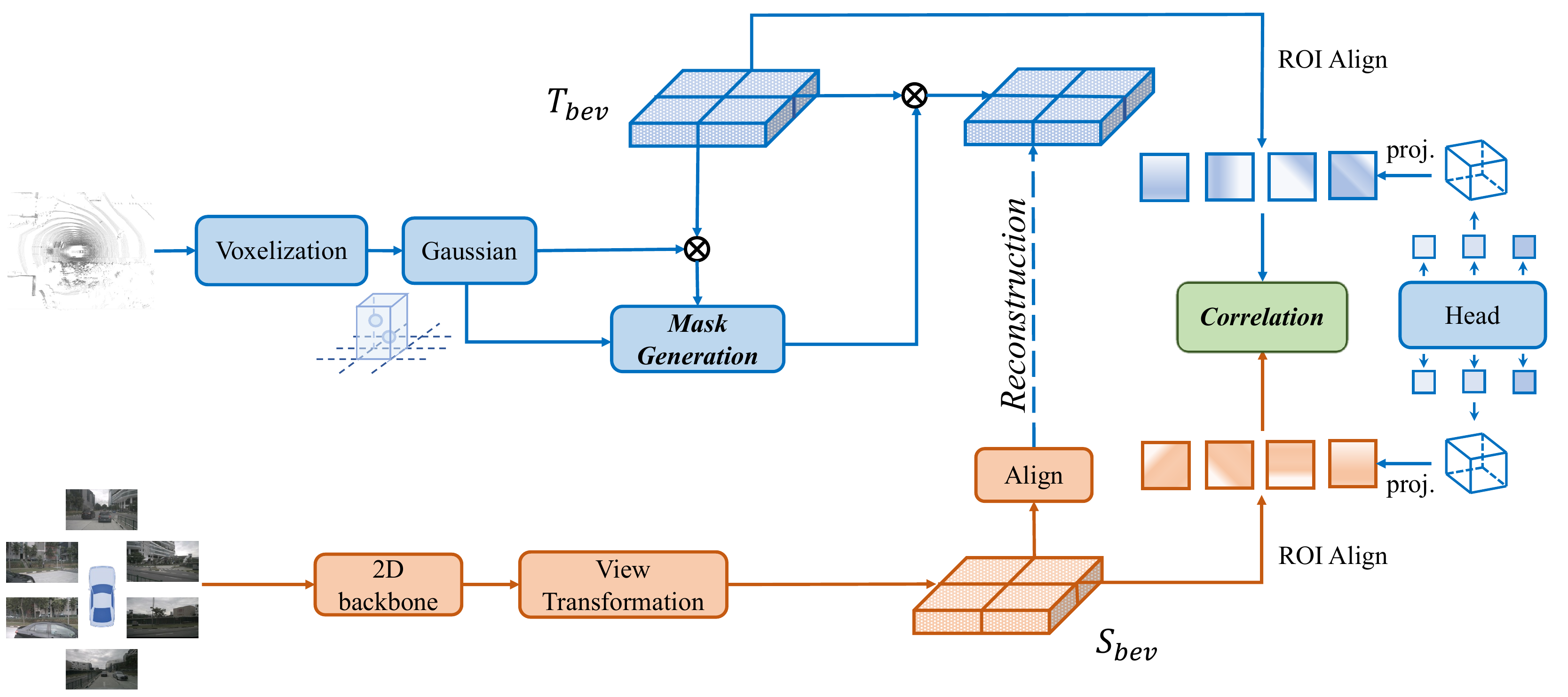}
    \caption{
    \textbf{Architecture of the GAPretrain.} The proposed \textbf{GAPretrain} is formulated as a training framework for camera-based methods.
    As indicated by blue arrows, we leverage geometric information from both point cloud data and LiDAR BEV representation in the pretraining phase.
    The goal is to learn spatial knowledge for the camera network from its LiDAR counterpart.
    $T_{bev}$ represents the BEV feature map generated by the LiDAR backbone. To make the figure more clear, we do not plot the generated process of the LiDAR BEV feature map $T_{bev}$. $S_{bev}$ indicates the BEV representation of the camera input.
    }
    \label{fig:overview}
\end{figure*}

Images captured by standard cameras only contain visual information flattened onto the 2D plane, 
making problems such as depth estimation, which require 3D information, naturally ill-posed.
Due to the lack of spatial and structural cues, 3D detection methods taking RGB images as input hardly provide satisfactory perception results.
On the contrary,
benefiting from accurate spatial information of point cloud data, LiDAR-based methods achieve ideal performance. 
To bridge the performance gap between LiDAR- and camera-based approaches, we propose a general pretraining method, which utilizes geometry-rich information from another modality to guide the training process of camera models.
Our proposed framework is a general plug-and-play module that can be flexibly applied to multiple camera-based detectors.

\subsection{Unified BEV Representation}

Data collected by different sensors exist in different views. 
Cameras take images in the perspective view and only cover a limited field of view (FOV), 
while LiDARs capture the complete scene in a frame of point cloud.
This view discrepancy makes knowledge transfer across different sensors non-trivial.
%
State-of-the-art multi-camera methods usually transform image features into dense BEV representation, on which the task of 3D object detection is performed.
Meanwhile, point cloud data can be easily transformed into BEV by flattening the height dimension.
Thus, features in the BEV space can be regarded as the bridge between different modalities and provide a unified representation.
Without loss of generality, our method can be applied to multi-camera detection methods with an explicit BEV feature representation.

Let $P \in \mathbb{R}^{N \!\times\! (3+K)}$ denotes a point cloud captured in a scene, where $N$ is the number of points and $K$ is the number of additional features except 3D coordinate.
A 3D sparse convolutional network consumes a point cloud and outputs a feature vector for each non-empty voxel.
As standard detection heads take dense 2D tensors as input, the sparse voxel representations are converted into BEV feature maps by reducing the height dimension.
We use $b_{L}$ to summarize the process of producing an output BEV representation from an input point cloud.
%
In the same scene, multiple synchronized images are captured to cover the entire horizontal FOV of the surrounding environment. 
Multi-view RGB images taken by $M$ cameras are denoted as $I \in \mathbb{R}^{M \!\times\! 3 \!\times\! H \!\times\! W}$,
where the image resolution is $H \times W$.
Images in perspective view are first processed by a typical 2D backbone to conduct feature extraction.
As the spatial relationship between the Lidar and cameras is known, the process of view transformation can be applied to generate BEV representation from image features with additional 3D prior.
We abstract the image backbone and the view transformation module as $b_C$.
The shape of the two BEV representations is set to be the same that $b_L(P), b_C(I) \in \mathbb{R}^{D \!\times\! X \!\times\! Y}$, where $D$ is the number of channels, and the spatial shape is defined as $X$ and $Y$.

\subsection{LiDAR-to-camera Pretraining}

Given the unified BEV representation on which object detection is performed, the geometry-aware BEV features from the LiDAR network serve as ideal pretraining targets.
To provide representative features, the LiDAR model $h_L \circ b_L$ is first trained on the 3D object detection task, 
where $h_{L}$ is the detection head.
The learned backbone and detection head are denoted as $\tilde{b}_{L}$ and $\tilde{h}_{L}$ respectively.
As the LiDAR model is only supervised by the detection signals, the BEV feature is an intermediate representation and not explicitly constrained.
It leads to a result that the value magnitudes of different channels distribute arbitrarily, which harms the optimization of the camera model.
To solve this problem, we regularize the BEV feature map with a normalization operation, which is implemented by calculating the channel-wise statistics of all training data.

The process of view transformation can be categorized into 2D-3D and 3D-2D methods.
As both types of view transformation methods contain learnable parameters,
the BEV features $b_C(I)$ inevitably become noisy due to imperfect conversion.
In contrast, the LiDAR backbone preserves accurate spatial information by maintaining the 3D coordinate of each point.
Due to the sparse nature of point cloud data, the resulting BEV features also contain empty regions.
For cross-modal knowledge transfer,
forcing BEV representations from LiDAR and camera to be aligned in these mismatched areas would raise confusion and bring a negative impact because of the enormous difference in between.

In the task of 2D object detection, methods of knowledge distillation endeavor to distinguish important foreground features.
This similar measure can be applied in our case to focus knowledge transfer exclusively on foreground objects, as the BEV features in these locations are usually dense.
However, this does not suffice the 3D detection task.
The reason might be that localization information is the mainstay of object detection in the BEV space.
As the range of BEV features reaches a distance of up to a hundred meters, the local foreground features for objects fail to determine the exact locations.
Background information, such as the type of ground and the surrounding static traffic signs, also plays an essential role.

\textbf{LiDAR-guided Mask Generation.}
Therefore, we leverage the spatial hints from point cloud data as additional guidance to emphasize foreground instances and background information.
Denoting the perceived BEV space as $W$, we divide it into $M \times N$ uniform grids, and the grid with the index $(i, j)$ is written as $w_{i,j}$. Given collected LiDAR points $P_{l} \in \mathbb{R}^{N_{p} \times 3}$ and synchronized camera images, we project the LiDAR point cloud to the grids in $W$ and count the number of point within each grid.
Specifically, we calculate statistics from raw point clouds and utilize them as a continuous attentive mask  $W \in \mathbb{R}^{X \!\times\! Y}$ with the same spatial shape as the BEV feature. To solve the misalignment between LiDAR points and the images, we adopt the Gaussian smoothing kernel to densify the LiDAR attention map. Assuming the LiDAR point number within the grid $w_{i,j}$ is $m_{i,j}$, the LiDAR attention map is represented as follows:
\begin{equation}
    \widetilde{m}_{i,j} = \sum_{i=-1}^{1}\sum_{j=-1}^{1} G_{i,j} \cdot m_{i,j},
\end{equation}
where $G$ represents Gaussian alignment that a Gaussian smoothing kernel of shape $(3,3)$.
Besides, some points are not required by camera-based models. To sample the valuable and aligned points used for knowledge transfer, we design a module to take advantage of the LiDAR BEV knowledge used to sample the important points. As the LiDAR BEV features are aligned with the Ground Truth, we choose it as the $Prior$ in the sampling process:
\begin{equation}
    c = \widetilde{m}(u, v) \cdot F^l(u, v),
\end{equation}
where $F^{l}$ represents the LiDAR BEV features. By simply setting a threshold $\theta$, we can sample the high respondence area $\widetilde{c}$ to transfer the aligned knowledge:

\begin{equation}
    \widetilde{m}_{i,j} = \widetilde{m}_{i,j} \cdot \widetilde{c},
\end{equation}
\begin{equation}\label{Eq1}
    R_{i,j} = \sqrt{ \text{ln} \{\widetilde{m}_{i,j} + 1\}} , 
\end{equation}
where $R_{i,j}$ denotes the $(i,j)$ element in $R$. 

We observe that two kinds of regions in $\widetilde{m}_{i,j}$ obsess larger value, \textit{i.e.}, the regions with more LiDAR points highlighted in $\widetilde{F}^l$. Among them, the area with a large amount of LiDAR points implies that there exist obstacles close to the ego vehicle. The areas highlighted in $\widetilde{F}^l$ indicate that these areas are important for the detection process of the LiDAR-based detector. 

As a result, we can not only highlight the important area but also rectify the location projected from the perspective view into the BEV. Meanwhile, it can also alleviate the depth estimation problem by learning geometry and depth information after view transformation. 
In short, $R_{i,j}$ describes which areas of representation in $W$ (could be the background regions in BEV) correspond to the right part in perspective view.

Then $\mathcal{L}_2$ norm loss is applied in the reconstruction process:
\begin{equation} \label{Eq2}
    \mathcal{L}_{rec} = \frac{1}{X \cdot Y} \sum_i^X \sum_j^Y R_{ij} \big|\big| \tilde{b}_L(P)_{ij} - b_C(I)_{ij} \big|\big|^2.
\end{equation}

\textbf{Target-aware Geometry Correlation.}
As the pixel-to-pixel knowledge transfer is difficult between two different modalities. To this end, we propose a different scheme termed Target-aware Geometry Correlation to solve this problem. Specifically, we extract the instance features by utilizing the coordinate of the prediction $\hat{p}(x,y,z,w,h,l,\theta)$ in the LiDAR model. To construct a new box without orientation $\theta$, we begin by projecting the coordinates $\hat{p}$ in 3D space onto a 2D plane and then acquiring the boxes $c(x_i, y_i)$. This enables us to create the extracted boxes $\tilde{c}(\tilde{x_i},\tilde{y_i})$ without specified orientation:
\begin{equation}\label{Eq3}
    \tilde{x}_{tl} = \mathop{\text{min}}\limits_{i \in \{0,1,2,3\}} x_i , 
\end{equation}
\begin{equation}\label{Eq4}
    \tilde{y}_{tl} = \mathop{\text{max}}\limits_{i \in \{0,1,2,3\}} y_i , 
\end{equation}
where $i$ indexes the corners of the projected boxes $c$. $\tilde{x}_{tl}$ and $\tilde{y}_{tl}$ denotes the top left corners of the extracted boxes $\tilde{c}$.

The instance features are extracted and resized to the same size:
\begin{equation}\label{Eq5}
    \hat{x} = \psi(f, c, o) , 
\end{equation}
where $\psi$ represents ROI Align~\citep{he2017mask}. $f$ is the input feature map. o indicates the size of the output feature:
\begin{equation}\label{Eq6}
    \mathcal{L}_{corr} = \frac{1}{N}\sum_{k=1}^N \text{SmoothL1Loss}(\phi_1(\hat{x}_{sk}), \phi_2(\hat{x}_{tk})), 
\end{equation}
where $\phi$ is a MLP network to extract the correspondence geometry information from the feature map. $\hat{x}_{s}$ denotes the extracted feature from the camera-based BEV feature map, while $\hat{x}_{t}$ represents the extracted feature from the LiDAR-based BEV feature map. $N$ depicts the number of the extracted feature map. To prevent the model from collapsing, $\phi_2$ is fixed during the pretraining stage. 

\textbf{Overall Loss Function.}  
According to the analysis above, the overall loss function consists of the following two parts. The $\mathcal{L}_{rec}$ is the pixel-wise reconstruction loss with the generated weight. $\mathcal{L}_{corr}$ is the correlation between the cropped area. The overall loss function is represented as:
\begin{equation}
    \mathcal{L}_{pretrain} = \lambda_1 \cdot \mathcal{L}_{rec} + \lambda_2 \cdot \mathcal{L}_{corr},
\end{equation}
where $\lambda_1$ and $\lambda_2$ are hyper-parameters to balance the scale of each loss.

\subsection{Finetuning}
In the finetuning stage, we follow the default settings of the different models. This enhances the performance of 3D detection models by leveraging improved initialization parameters. Specifically, we reuse these parameters directly in the finetuning phase, employing only images as input and not requiring LiDAR points. This strategy reduces the computational burden and eliminates the need for hyperparameter tuning to improve the accuracy of 3D detection models.
During the pretraining phase, the optimization process of different detection heads yields a diverse bird's-eye view (BEV) representation. To align the BEV representation between LiDAR and camera models, we design the camera model's detection head architecture to be identical to that of the LiDAR model. This approach better leverages consistent BEV features and leads to improved performance.
To ensure that the learned BEV representation $F^c$ of the camera model follows the same distribution as $F^l$ after the pretraining stage, we utilize the LiDAR head parameters in the finetuning phase. This technique simplifies the optimization process and results in better performance. We demonstrate the effectiveness of our approach through experimentation and validation.

\begin{table*}[ht!]
    \caption{
    \textbf{Numeric results for 3D object detection task on nuScenes \textit{val} set. }
    $^\dag$ indicates we utilize additional unlabeled data.
    Our approach unanimously improves performance on various backbone networks and view transformation. 
    The same conclusion applies to models with temporal information.
    }
    \label{tab:baselines}
    \vskip 0.15in
    \begin{center}
    \resizebox{\linewidth}{!}{
        \begin{tabular}{c|cccc|cc|ccccc}
            \toprule
            Method & Backbone & View Transformation & Temporal & Ours & mAP$\uparrow$ & NDS$\uparrow$ & mATE$\downarrow$ & mASE$\downarrow$ & mAOE$\downarrow$ & mAVE$\downarrow$ & mAAE$\downarrow$ \\
            \midrule
            \multirow{3}{*}{BEVFusion-C~\citep{liang2022bevfusion}} & Dual-Swin-Tiny & LSS &  - &  & 34.4 & 35.8 & 0.703 & 0.296 & 0.740 & 1.249 & 0.400 \\
             &  \gray Dual-Swin-Tiny  & \gray LSS & \gray - & \gray\cmark &  \gray38.0 \textit{\color{red}{(+3.6)}} & \gray42.6 \textit{\color{red}{(+6.8)}} & \gray0.661 & \gray0.268 & \gray0.554 & \gray0.937 & \gray0.220 \\
             &  \gray Dual-Swin-Tiny  & \gray LSS & \gray - & \gray\cmark$^\dag$ &  \gray38.8 \textit{\color{red}{(+4.4)}} & \gray44.2 \textit{\color{red}{(+8.4)}} & \gray0.632 & \gray0.270 & \gray0.524 & \gray0.873 & \gray0.218 \\
            \midrule
            \multirow{2}{*}{BEVDepth~\citep{li2022bevdepth}} & ResNet-50  & LSS & Concatenate && 36.4 & 45.4 & 0.707 & 0.283 & 0.655 & 0.418 & 0.216 \\
            & \gray ResNet-50 & \gray LSS & \gray Concatenate &  \gray\cmark & \gray38.2 \textit{\color{red}{(+1.8)}} & \gray49.0 \textit{\color{red}{(+3.6)}} & \gray0.670 & \gray0.275  & \gray0.500 & \gray0.366  &  \gray0.198 \\
            \midrule
            \multirow{6}{*}{BEVFormer~\citep{li2022bevformer}} & ResNet-18  & Deformable  & Recurrent & & 26.2 & 38.3 & 0.853 & 0.289 & 0.612 & 0.511 & 0.218 \\
             & \gray ResNet-18  & \gray Deformable  &  \gray Recurrent & \gray\cmark & \gray33.1 \textit{\color{red}{(+6.9)}} & \gray44.9 \textit{\color{red}{(+6.6)}} & \gray0.721 & \gray0.281 & \gray0.543 & \gray0.413 & \gray0.202 \\
             & ResNet-34 & Deformable  & Recurrent &  & 31.8  & 43.3 & 0.784 & 0.285 & 0.513 & 0.474 & 0.201\\
             & \gray ResNet-34 & \gray Deformable  & \gray Recurrent & \gray\cmark & \gray37.4 \textit{\color{red}{(+5.6)}} & \gray48.3 \textit{\color{red}{(+5.0)}} & \gray0.690 & \gray0.279 & \gray0.457 & \gray0.414 & \gray0.201 \\
             & ResNet-50  & Deformable  & Recurrent  && 33.6  & 45.2 & 0.757 & 0.282 & 0.477 & 0.443 & 0.204 \\
             & \gray ResNet-50 & \gray Deformable  & \gray Recurrent  & \gray\cmark & \gray38.4 \textit{\color{red}{(+4.8)}} & \gray49.1 \textit{\color{red}{(+3.9)}} & \gray0.684 & \gray0.274 & \gray0.441 & \gray0.412 & \gray0.196  \\
            \bottomrule
        \end{tabular}
    }
    \end{center}
    \vskip -0.1in
\end{table*}


\section{Experiments}

In this section, we first introduce our experimental setups, including dataset, metrics, and implementation details. 
Then experiments on various multi-camera methods are conducted to validate the effectiveness of our proposed pipeline.

\subsection{Dataset and Metrics}

The nuScenes dataset~\citep{caesar2020nuscenes} is a large-scale autonomous driving dataset containing 1000 driving scenes, in which 700, 150, and 150 are for training, validation, and testing respectively.
Each scene lasts about 20 seconds and is sampled at 2Hz for annotation.
Each frame consists of RGB images with a resolution of $900 \!\times\! 1600$ from 6 cameras covering the entire horizontal FOV.

Two primary metrics are provided for the 3D detection task: mean average precision (mAP) and nuScenes detection score (NDS).
The mAP is computed using the center distance on the ground plane to match the predicted results and ground truths. 
The nuScenes dataset defines five types of true positive metrics, namely mean Average Translation Error (mATE), mean Average Scale Error (mASE), mean Average Orientation Error (mAOE), mean Average Velocity Error (mAVE), and mean Average Attribute Error (mAAE), to measure translation, scale, orientation, velocity, and attribute errors respectively. 
The NDS is defined as a weighted sum of mAP and five true positive metrics.

\begin{table*}
    \caption{
        \textbf{Performance compared with state-of-the-art models} for the task of 3D object detection on nuScenes \textit{val} set.
        $^\ddagger$ means that ResNet-101 is initialized with the checkpoint from FCOS3D~\citep{wang2021fcos3d}. 
        * denotes that VoVNet-99 is pretrained on the depth estimation task with extra data~\citep{park2021dd3d}. 
        With VoVNet-99 as the image backbone and pretrained by the proposed method, BEVFormer achieves state-of-the-art performance.
    }
    \label{comparision_val}
    \begin{center}
    \resizebox{\linewidth}{!}{
        \begin{tabular}{c|c|c|cc|ccccc}
            \toprule
            Method & Backbone  & epoch & mAP$\uparrow$ & NDS$\uparrow$ & mATE$\downarrow$ & mASE$\downarrow$ & mAOE$\downarrow$ & mAVE$\downarrow$ & mAAE$\downarrow$ \\
            \midrule
            FCOS3D~\citep{wang2021fcos3d} & ResNet-101 & 24 & 29.5 & 37.2 & 0.806 & 0.268 & 0.511 & 1.315 & 0.199 \\
            DETR3D~\citep{wang2022detr3d} & ResNet-101$^\ddagger$ &  24 & 34.6  & 42.5 & 0.773 & 0.268 & 0.383 & 0.842 & 0.216\\
            PETR~\citep{liu2022petr} & ResNet-101$^\ddagger$  & 24 & 37.0  & 44.2 & 0.711 & 0.267 & 0.383 & 0.865 & 0.201\\
            UVTR~\citep{li2022uvtr} & ResNet-101$^\ddagger$ &  24 & 37.9  & 48.3 & 0.731 & 0.267 & 0.350 & 0.510 & 0.200\\
            BEVDet ~\citep{huang2021bevdet} & Swin-B & 90 & 39.3  & 47.2 & 0.608 & 0.259 & 0.366 & 0.822 & 0.191\\
            BEVDepth~\citep{li2022bevdepth} & ResNet-101 &  90 & 41.2  & 53.5 & 0.565 & 0.266 & 0.358 & 0.331 & 0.190\\
            BEVDet4D~\citep{huang2022bevdet4d} & Swin-B & 90 & 42.1  & 54.5 & 0.579 & 0.258 & 0.329 & 0.301 & 0.191\\
            PETRv2~\citep{liu2022petrv2} & ResNet-101$^\ddagger$ &  24 & 42.1  & 52.4 & 0.681 & 0.267 & 0.357 & 0.377 & 0.186\\
            STS~\citep{wang2022sts} & ResNet-101 &  90 & 43.1  & 54.2 & 0.525 & 0.262 & 0.380 & 0.369 & 0.204\\
            PolarFormer~\citep{jiang2022polarformer} & ResNet-101$^\ddagger$ &  24 & 43.2  & 52.8 & 0.648 & 0.270 & 0.348 & 0.409 & 0.201\\
            \midrule
            \multirow{4}{*}{\shortstack{BEVFormer}}
             & ResNet-101$^\ddagger$ & 24 & 41.5  & 51.7 & 0.672 & 0.274 & 0.369 & 0.397 & \textbf{0.198} \\
             & \gray ResNet-101$^\ddagger$ + ours & \gray24 &  \gray\textbf{44.0}\textit{\color{red}{(+2.5)}}  & \gray\textbf{54.2}\textit{\color{red}{(+2.5)}}  & \gray\textbf{0.639} & \gray\textbf{0.268} & \gray\textbf{0.331} & \gray\textbf{0.338} & \gray0.203  \\
             & VoVNet-99*  & 24 & 43.5 & 53.4 & 0.667 & 0.276 & \textbf{0.339} & 0.360 & 0.195 \\
             & \gray VoVNet-99* + ours  & \gray24 & \gray\textbf{46.2}\textit{\color{red}{(+2.7)}}  & \gray\textbf{55.5}\textit{\color{red}{(+2.1)}} & \gray\textbf{0.617} & \gray\textbf{0.264} & \gray0.357 & \gray\textbf{0.335} & \gray\textbf{0.185} \\
            \bottomrule
        \end{tabular}
    }
    \end{center}
\end{table*}

\subsection{Implementation Details}
\label{sec:implementation}

The input point clouds are filtered by the range of $[-51.2m, 51.2m]$, except that the range of BEVFusion-C follows the open-source implementation as $[-54m, 54m]$.
Since the BEV features of different camera models are in different shapes, the voxel sizes of LiDAR models are changed accordingly.
In the phase of pretraining, augmentation in both modalities is turned off to spatially align BEV representations.
Camera models are pretrained by 24 epochs with the AdamW optimizer.
To make a fair comparison to baseline models, settings in the phase of finetuning, such as data augmentations and training schedules, are kept the same for each camera network.
We utilize the BEVFormer, BEVDepth, BEVFusion as the baseline. For ablation experiments, we use the BEVFusion-C and BEVFormer as the frameworks to validate the effectiveness of our pretraining method.

\textbf{BEVFusion-C.}
BEVFusion-C~\citep{liang2022bevfusion} denotes the camera branch of the multi-model model.
It uses Dual-Swin-Tiny~\citep{liu2020dual} as the image backbone.
The classic LSS~\citep{philion2020lss} method is used for view transformation.
The BEV feature map is in the shape of $512 \!\times\! 180 \!\times\! 180$, where 512 is the number of channels and 180 is the spatial shape.
Input images have a resolution of $448 \!\times\! 800$. 
Note that this model does not introduce the temporal information, and thus is treated as a lower baseline.

\textbf{BEVDepth.} 
BEVDepth~\citep{li2022bevdepth} adopts ResNet-50 as its image backbone.
With the predicted depth information, efficient voxel pool transforms image features into BEV space.
It has a BEV representation in 256 channels and the shape of $128 \!\times\! 128$, and takes images with a resolution of $512 \!\times\! 1408$ as input.
Alignment of the pseudo point cloud from the previous timestep is utilized to introduce temporal information.

\textbf{BEVFormer.}
Following the same settings as the submitted version to the nuScenes leaderboard,
BEVFormer~\citep{li2022bevformer} employs VoVNet-99~\citep{lee2019vovnet} as the image backbone.
Besides, we also adopt extra light-weighted image backbones, namely ResNet-18 
and ResNet-50, to study the effect of model capacity.
The BEV feature is set to the size of $200 \!\times\! 200$ with 256 channels.
The image backbone consumes RGB images in the shape of $900 \!\times\! 1600$.
Temporal information is encoded in a recurrent manner.

\begin{table*}[!ht]
    \caption{
    \textbf{Comparision among different methods.}  To conduct the fair comparision, we adopt different methods to train on the BEVFormer-R50 for 36 epoch with LiDAR teacher. $^\dagger$ indicates our model is pretraining for 12 epoch, then fine-tune for 24 epoch without LiDAR models. $^*$ denotes our reimplementation of BEVDistill.
    }
    \label{tab:compare}
    \vskip 0.15in
    \begin{center}
    \resizebox{.95\linewidth}{!}{
        \begin{tabular}{c|c|c|cc|ccccc}
            \toprule
             Detector  & Setting & epoch & mAP$\uparrow$ & NDS$\uparrow$ &mATE$\downarrow$ & mASE$\downarrow$ & mAOE$\downarrow$ & mAVE$\downarrow$ & mAAE$\downarrow$  \\ 
            \midrule
              \multirow{7}{*}{\makecell[c]{BEVFormer-R50}}
               & \gray Teacher~\citep{bai2022transfusion} & \gray20 & \gray63.2  & \gray68.4 & \gray0.324 & \gray0.265 & \gray0.295 & \gray0.246 & \gray0.188\\
               & \gray Student~\citep{li2022bevformer} & \gray24 & \gray33.6  & \gray45.2 & \gray0.757 &\gray 0.282 & \gray0.477 & \gray0.443 & \gray0.204 \\
               & FitNet~\citep{romero2014fitnets} & 36 &  34.0 & 45.4 & 0.752 & 0.286 & 0.461 & 0.456 & 0.200   \\
               & CWD~\citep{shu2021channel} & 36 & 33.9 & 44.9 & 0.758 & 0.281 & 0.546 & 0.431 & 0.194 \\
               & UVTR~\citep{li2022uvtr} & 36 & 35.4 & 45.6 & 0.748 & 0.277 & 0.527 & 0.455 & 0.205 \\
               & BEVDistill$^*$~\citep{chen2022bevdistill} & 36 & 37.0 & 47.2 & 0.705 & 0.281 & 0.523 & 0.406 & 0.198 \\               
               & Ours & 36$^\dagger$ & 38.4\textit{\color{red}{(+4.8)}} & 49.1\textit{\color{red}{(+3.9)}} & 0.684 & 0.274 & 0.441 & 0.412 & 0.196 \\
            \bottomrule
        \end{tabular}
    }
    \end{center}
    \vskip -0.1in
\end{table*}

\subsection{Building on Top of SOTA}

We adopt the LiDAR backbone from TransFusion-L~\citep{bai2022transfusion}, which is a popular and efficient architectural design.
To prove the generalization ability of our proposed pipeline, we consider multiple models under different settings,
such as various types of image backbone, view transformation, and temporal information.

\cref{tab:baselines} describes the numeric results.
The proposed pipeline brings performance improvement to various state-of-the-art methods on nuScense \textit{val} set, validating the generalization ability of our approach.
BEVFusion-C has a significant improvement, with a gain of 3.6\% mAP and 6.8\% NDS respectively. Besides, model performance can be further boosted by utilizing unlabeled data in the pretraining
stage (see \cref{datascale} for details). With the help of the 100\% unlabeled data, BEVFusion-C improves the performance by 0.8\% mAP and 1.6\% NDS.
Without the use of temporal information, BEVFusion-C has limited performance on various attributes, particularly on the mAVE.
As the LiDAR network consumes ten consecutive frames as input, the implicit temporal information benefits the model performance.
By utilizing our pretraining method, both spatial and temporal knowledge is effectively injected into the camera model. Note that BEVDepth utilizes explicit depth supervision and temporal information, the proposed method can achieve  1.8\% mAP and 3.6\% NDS improvement upon it nonetheless.

\begin{table*}[th!]
    \caption{\textbf{The comparison among different ratios of pretraining data.} Pretrained with different percentages of data, model performance is shifted correspondingly.}
    \centering
    \resizebox{\linewidth}{!}{
        \scale
        \begin{tabular}{c|cc|ccccc}
            \toprule
            Data Scale & mAP$\uparrow$ & NDS$\uparrow$ & mATE$\downarrow$ & mASE$\downarrow$ & mAOE$\downarrow$ & mAVE$\downarrow$ & mAAE$\downarrow$ \\
            \midrule
            0\% & 34.4 & 35.8 & 0.703 & 0.296 & 0.740 & 1.249 & 0.400 \\
            10\% & 33.9 & 36.3 & 0.716 & 0.280 & 0.764 & 1.072 & 0.302 \\
            50\% & 35.8 & 39.4 & 0.683 & 0.274 & 0.671 & 0.981 & 0.243 \\
            100\% & 38.0 & 42.6 & 0.661 & 0.268 & 0.554 & 0.939 & 0.220 \\
            200\% & 38.8 & 44.2 & 0.632 & 0.270 & 0.524 & 0.873 & 0.218 \\
            \bottomrule
        \end{tabular}
    }
    \label{datascale}
\end{table*}

For BEVFormer, we adopt various image backbones with different network depths. All backbones can benefit from our pretraining method by a large margin. The metric mATE measures the Euclidean center distance between prediction and ground truth. Such object translation error is highly correlated with depth estimation. Therefore, the accuracy of depth estimation can be reflected by the mATE. As shown in \cref{tab:baselines}, for all methods, mATEs are improved significantly, which proves that our pretraining method greatly improves depth estimation performance of the model.

In \cref{comparision_val}, we compare the BEVFormer using our pretraining paradigm with the state-of-the-art methods. Notably, ResNet-101 with our pretraining method surpasses VoVNet-99 by a conspicuous margin, which demonstrates that our pretraining method can serve as a better parameter initialization for multi-view 3D object detection.
In addition, even though based on strong backbone VoVNet-99, which provides state-of-the-art results, we can still have a 2.7\% mAP and 2.1\% NDS gain. With our pretraining method, BEVFormer achieves 46.2\% mAP and 55.5\% NDS on nuScenes \textit{val} set, which surpasses other state-of-the-art results by a large margin.
Furthermore, the reported scores on nuScenes \textit{test} split are 49.1 mAP and 57.8 NDS, outperforming its original version by 1.0 mAP and 0.9 NDS.

\subsection{Ablation Study}

\subsubsection{Comparasion among Different Methods}
We implement BEVFormer-R50, a state-of-the-art deep neural network for camera-based 3D object detection, on the nuScenes dataset and compare the effectiveness of our proposed knowledge distillation method (GAPretrain) with that of other state-of-the-art methods. To conduct a fair comparison, we adopt different methods to train BEVFormer-R50 for 36 epochs with a LiDAR teacher, while GAPretrain uses LiDAR point clouds in the pretraining stage for 12 epochs. In the finetuning stage, we follow the default settings set in various models. Our GAPretrain method outperforms the vanilla student model by a large margin, achieving improvements of over 4.8 mAP and 3.9 NDS, respectively. Notably, end-to-end distillation methods such as FitNet~\citep{romero2014fitnets} and CWD~\citep{shu2021channel} fail to bring prominent improvement, likely due to their limited ability to capture the complex relationship between the two different modalities. UVTR, as introduced in \cite{li2022uvtr}, results in a 1.8 mAP improvement; however, it does not lead to a significant improvement in NDS. In contrast, GAPretrain manages to exceed the state-of-the-art distillation method, namely BEVDistill~\citep{chen2022bevdistill}, although BEVDistill underwent 36 epochs of training with LiDAR guidance. This is likely due to the fact that the end-to-end distillation manner is suboptimal. By incorporating a geometric-aware approach with cross-modal contrastive learning, GAPretrain is able to successfully address this limitation and improve the accuracy of the student model.

\begin{table*}[!ht]
    \caption{
    \textbf{Performance of the main components.} The task of 3D object detection demands not only pixel-wise but also instance-level correlation. The Pretraining distillation and Mask generation surpass the baseline by a large margin. The abbreviation ``TGC'' represents ``Target-aware Geometry Correlation''.
    The non-trivial performance gains is acquired after highlighting the target area with sparse supervision. 
    }
    \label{tab:component}
    \vskip 0.15in
    \begin{center}
    \resizebox{.95\linewidth}{!}{
        \begin{tabular}{ccc|cc|ccccc}
            \toprule
             Pretraining Distillation & Mask Generation & TGC & mAP$\uparrow$ & NDS$\uparrow$  &mATE$\downarrow$ & 
            mASE$\downarrow$ & mAOE$\downarrow$ & mAVE$\downarrow$ & mAAE$\downarrow$ \\ 
            \midrule
               &  &  & 33.6  & 45.2 & 0.757 & 0.282 & 0.477 & 0.443 & 0.204\\
             \cmark  &  &  & 36.0 & 46.4 & 0.743 & 0.277 & 0.491 & 0.459 & \textbf{0.189}\\
              \cmark & \cmark &  &  38.1 & 48.7 & 0.693 & 0.276 & 0.456 & 0.414 & 0.200 \\
              \cmark  & \cmark  & \cmark & \textbf{38.4} & \textbf{49.1} & \textbf{0.684} & \textbf{0.274} & \textbf{0.441} & \textbf{0.412} & 0.196 \\
            \bottomrule
        \end{tabular}
    }
    \end{center}
    \vskip -0.1in
\end{table*}

\subsubsection{Data Scale}

Data annotation demands expensive human efforts.
For public datasets on 3D perception tasks, only a tiny fraction of data is labeled. 
As data is partially annotated, unlabeled data usually remains unused for the fully-supervised training process.
As shown in \cref{datascale}, with the proposed pretraining method, model performance can be further boosted by utilizing unlabeled data in the pretraining stage.
We study the model performance using 0\%, 10\%, 50\%, 100\%, and 200\% of data.
0\% denotes that the model is trained from scratch in the finetuning stage, and 200\% denotes that the same amount of unlabeled data as labeled data are collected as input in the pretraining phase.

\begin{table*}[ht!]
    \caption{\textbf{Design choices.} Ablation study on nuScenes \textit{val} set with BEVFusion-C as camera model. It is non-trivial to perform cross-modal pretraining on camera-based methods directly, as the devil lies in these critical design choices.}
    \label{tab:ablations}
    \begin{center}
    \subfloat[
     As the scale of statistical points is large, thus adopting regularization is necessary. We find log function achieves better results.
        \label{tab:weight_scaling}
    ]
    {
        \begin{minipage}[b]{0.4\linewidth}{
            \begin{center}
                \tablestyle
                \begin{tabular}{x{55}|x{30}x{30}}
                    \toprule
                      \multirow{2}{*}{\makecell[c]{Regularization\\ Function}} & \multirow{2}{*}{mAP$\uparrow$} & \multirow{2}{*}{NDS$\uparrow$} \\   
                    &&\\
                    \midrule
                     - & 35.5 & 40.9 \\
                     $sigmoid$ & 38.1 & 42.5 \\
                     $log$  & \textbf{38.6} & \textbf{43.4} \\
                    \bottomrule
                    \multicolumn{3}{c}{~} \\
                \end{tabular}
            \end{center}
        }\end{minipage}
    }
    \hspace{1em} 
    \subfloat[
        Different from the common knowledge transfer settings such as $KL$ loss, $\mathcal{L}_2$ benefits the pretraining most from three losses.
        \label{tab:loss}
    ]{  
        \begin{minipage}[b]{0.4\linewidth}{
            \begin{center}
                \tablestyle
                \begin{tabular}{x{55}|x{30}x{30}}
                    \toprule
                     \multirow{2}{*}{\makecell[c]{Pretraining \\ Loss Type}} & \multirow{2}{*}{mAP$\uparrow$} & \multirow{2}{*}{NDS$\uparrow$} \\  
                     &&\\
                    \midrule
                      $KL$  & 37.6 & 41.0 \\
                      $\mathcal{L}_1$ & 33.4 & 34.8 \\
                      $\mathcal{L}_2$  & \textbf{38.6} & \textbf{43.4} \\
                    \bottomrule
                    \multicolumn{3}{c}{~} \\
                \end{tabular}
            \end{center}
        }\end{minipage}
    }
    \vspace{0.005em} 
    \subfloat[
        By softening the magnitudes, normalized BEV features serve as ideal pretraining targets compared to the raw representations.
        \label{tab:normalization}
    ]{  
        \begin{minipage}[b]{0.4\linewidth}{
            \begin{center}
                \tablestyle
                \begin{tabular}{x{55}|x{30}x{30}}
                    \toprule
                     \multirow{2}{*}{\makecell[c]{Feature\\ Whitening}} & \multirow{2}{*}{mAP$\uparrow$} & \multirow{2}{*}{NDS$\uparrow$} \\
                     &&\\
                    \midrule
                      & 36.1 & 39.1 \\
                      \cmark  &  \textbf{38.6} & \textbf{43.4} \\
                    \bottomrule
                    \multicolumn{3}{c}{~} \\
                \end{tabular}
            \end{center}
        }\end{minipage}
    }
    \hspace{1em} 
    \subfloat[
        Trained along with the LiDAR network, the detection head benefits from the learned weights in the fine-tuning process.
        \label{tab:duplicate}
    ]{  
        \begin{minipage}[b]{0.4\linewidth}{
            \begin{center}
                \tablestyle
                \begin{tabular}{x{55}|x{30}x{30}}
                    \toprule
                     \multirow{2}{*}{\makecell[c]{Head\\ Inheritance}} & \multirow{2}{*}{mAP$\uparrow$} & \multirow{2}{*}{NDS$\uparrow$} \\
                     && \\
                    \midrule
                      & 38.0  & 41.1 \\
                      \cmark  &  \textbf{38.6} & \textbf{43.4} \\
                    \bottomrule
                    \multicolumn{3}{c}{~} \\
                \end{tabular}
            \end{center}
        }\end{minipage}
    }
    \end{center}
    \vskip -0.35in
\end{table*}

\subsubsection{The Main Component}
To confirm the contribution to the final results of each module, We perform ablation experiments on them. The main components include Pretraining Distillation, Mask Generation, and Target-aware Geometry Correlation. 

Pretraining Distillation divides the training process into two separate stages in order to mitigate the complexities involved in optimization. By jointly pretraining image backbone and view transformation, vision-centric object detectors learn more accurate spatial information and thus result in enhanced model performance. Furthermore, pretraining distillation can benefit from a large amount of unlabeled data.

Mask Generation combines the LiDAR representation with the processed LiDAR-guided attention map to compensate for the weakness of the sparse 3D box annotations. Gaussian operation spreads the points around objects using the Gaussian kernel of shape (3,3) to alleviate quantization error. By masking low-response pixels, we can significantly decrease the domain gap between LiDAR and the image.

The module of Target-aware Geometry Correlation extracts geometry information from the area predicted by the LiDAR model. It can reduce the domain gap by concentrating more on the whole correlation between different instances instead of pixel-wise supervision. 

As demonstrated in \cref{tab:component}, the Pretraining Distillation improves the performance of the model by 2.4\% mAP, even only with the vanilla feature reconstruction. Moreover, with the help of Mask Generation module, we acquire the more accurate BEV feature, improving the object localization (mATE) by 5.9\%. By highlighting the correlation between instances, we also get a further 0.4\% NDS performance gain.

\begin{figure*}[h!]
    \centering
    \includegraphics[width=.9\linewidth]{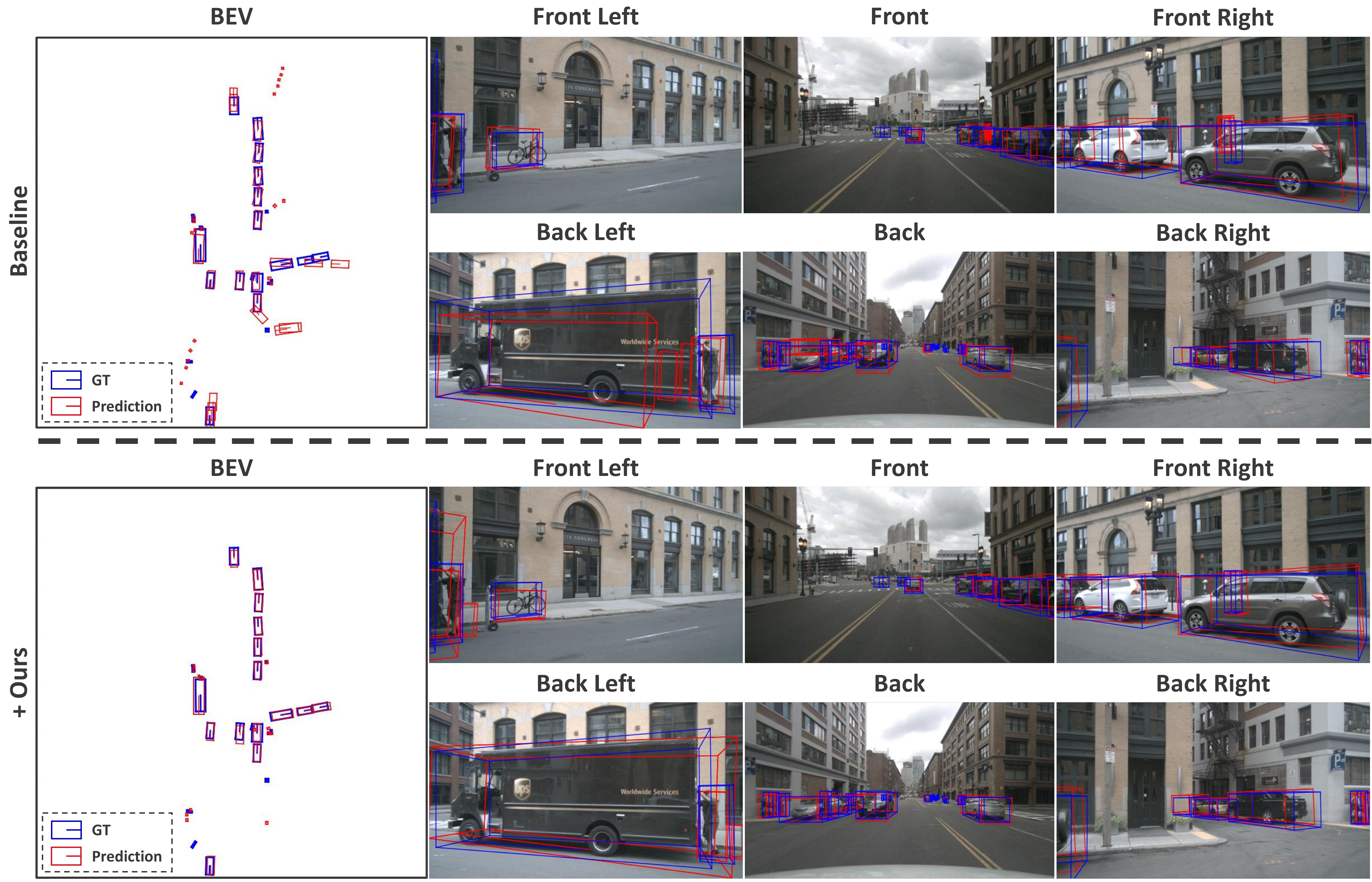}
    \caption{
    \textbf{Comparison between baseline and our method.}
     As depicted in the BEV, the predicted locations of many objects by the baseline
model shift from their real sites. Pretrained on geometric-rich targets, predictions of ours are more accurate.
    }
    \label{compare}
\end{figure*}

\subsubsection{Other Design Choices}

We conduct various experiments based on BEVFusion-C to validate the effectiveness of critical design choices in pretraining.

\textbf{Regularization Function.}
As the statistical LiDAR points are relatively large during the generation process of the LiDAR-guided attention map, we adopt different regularization functions to scale it. As pointed out in \cref{tab:weight_scaling}, the log function achieves the best results compared with others. Notably, the log function achieved a 3.1\% mAP performance gain compared with the baseline.

\textbf{Pretraining Loss Type.}
Our default pretraining loss is $L2$ loss. To better understand the effectiveness of different pretraining loss, we conduct experiments as noted in \cref{tab:loss}. It demonstrates that $L2$ loss achieves the best results. Moreover, this experiment also proves that our pretraining phase only needs the simple $L2$ loss rather than KL loss which is usually used in the knowledge transfer setting.

\textbf{Feature Whitening.}
We ablate the importance of the feature whitening.
We disable learnable scale and bias for BatchNorm to avoid collapse. Specifically, instead of directly adopting BatchNorm, we calculate the mean and variance of the whole dataset as the fixed number.
The experiments demonstrate the effectiveness of our normalization as pointed out in \cref{tab:normalization}. Instead of directly mimicking the LiDAR BEV feature, a channel-wise normalization is conducted on BEV feature maps from the LiDAR backbone.

\textbf{Head Inheritance.}
Head Inheritance is a common practice in single-modal distillation~\citep{kang2021headinherit}. However, we are exploring a more challenging setting with cross-modal representations. We directly compare head inheriting with the baseline as pointed in \cref{tab:duplicate}. Credit to a better initialization of the detection head, an improvement of 0.6\% and 2.3\% of mAP and NDS can be observed.

\subsection{Visualization}

We present BEV representations of models trained with and without our method in \cref{compare}.
As shown in the results, the proposed method refines the locations of predicted results and removes some false positive predictions.
Visually, the ray-like effect and noise in the BEV feature map from the baseline model are mitigated, as the object locations can be easily observed.


\section{Conclusion}

In this work, we propose GAPretrain, an effective training strategy for multi-camera methods on the task of 3D object detection. 
With its simplicity, our method serves as a plug-and-play approach for multi-view camera models with an explicit BEV representation, despite the diverse architectural designs of the image backbone and view transformation.
Benefiting from the pretraining process, state-of-the-art camera-based methods gain further performance improvement.
As LiDAR and camera data without annotation can serve as the input in the pretraining phase, the massive unlabeled data can be utilized.
We hope this work can serve as a baseline for the pretraining of camera-based 3D detection methods.

As a learned neural network, the LiDAR model provides imperfect pretraining objectives in the BEV space.
Moreover, due to the nature of the sparsity, BEV representations from LiDAR point cloud could not benefit the performance on objects at a large distance.
Future works could focus on producing more representative and robust targets.

\section*{Data availability statement}\label{sec8}
The datasets analyzed in this study are accessible on \url{https://www.nuscenes.org/nuscenes}.


\bibliography{sn-article}

\end{document}